\documentclass[runningheads]{llncs}
\usepackage{graphicx}
\pdfoutput=1

\usepackage[hang,flushmargin]{footmisc}
\usepackage[misc,geometry]{ifsym}
\usepackage{cite} 
\usepackage[T1]{fontenc}
\usepackage{amsmath}
\usepackage{mathrsfs}
\usepackage{graphicx}  
\usepackage{epstopdf}
\usepackage{bbm}
\usepackage{float}
\usepackage{multirow}
\usepackage{booktabs}
\usepackage{mathrsfs}
\usepackage{amssymb}
\usepackage[colorlinks,linkcolor=red]{hyperref}
\usepackage{siunitx}

\usepackage{arydshln}

\makeatletter

\newcommand{\Rmnum}[1]{\expandafter\@slowromancap\romannumeral #1@}
\makeatother
\usepackage{booktabs}
\usepackage{amsfonts,amssymb}
\usepackage{amsmath,bm}
\usepackage{array}
\newcommand{\PreserveBackslash}[1]{\let\temp=\\#1\let\\=\temp}
\newcolumntype{C}[1]{>{\PreserveBackslash\centering}p{#1}}
\newcolumntype{R}[1]{>{\PreserveBackslash\raggedleft}p{#1}}
\newcolumntype{L}[1]{>{\PreserveBackslash\raggedright}p{#1}}
\usepackage{algorithmicx,algorithm}
\makeatletter

\newcommand{\para}[1]{\vspace{.05in}\noindent\textbf{#1}}
\def\ie{\emph{i.e.}}

\usepackage{xcolor}

\begin{document}
\title{HIGT: Hierarchical Interaction Graph-Transformer for Whole Slide Image Analysis}
%
\titlerunning{Hierarchical Interaction Graph-Transformer}
%
\authorrunning{Z. Guo et al.}
\author{
Ziyu Guo\inst{1}$^{*}$\and
Weiqin Zhao\inst{1}\thanks{Contributed equally to this work.}\and
Shujun Wang\inst{2}\and
Lequan Yu\inst{1}
}
\institute{
    The University of Hong Kong, Hong Kong SAR, China \and
    The Hong Kong Polytechnic University, Hong Kong SAR, China
}
    

%
%

\maketitle              

\begin{abstract}
In computation pathology, the pyramid structure of gigapixel Whole Slide Images (WSIs) has recently been studied for capturing various information from individual cell interactions to tissue microenvironments. 
This hierarchical structure is believed to be beneficial for cancer diagnosis and prognosis tasks. 
However, most previous hierarchical WSI analysis works 
(1) only characterize local or global correlations within the WSI pyramids
and (2) use only unidirectional interaction between different resolutions, leading to an incomplete picture of WSI pyramids.
To this end, this paper presents a novel Hierarchical Interaction Graph-Transformer (\ie, HIGT) for WSI analysis.
With Graph Neural Network and Transformer as the building commons, HIGT can learn both short-range local information and long-range global representation of the WSI pyramids. 
%
Considering that the information from different resolutions is complementary and can benefit each other during the learning process, we further design a novel Bidirectional Interaction block to establish communication between different levels within the WSI pyramids.
Finally, we aggregate both coarse-grained and fine-grained features learned from different levels together for slide-level prediction.
We evaluate our methods on two public WSI datasets from TCGA projects, \ie, kidney carcinoma (KICA) and esophageal carcinoma (ESCA). 
Experimental results show that our HIGT outperforms both hierarchical and non-hierarchical state-of-the-art methods on both tumor subtyping and staging tasks.

\keywords{ WSI analysis  \and Hierarchical representation \and Interaction \and Graph neural network \and Vision transformer}
\end{abstract}

\section{Introduction}
\begin{figure}[t]
\includegraphics[width=\textwidth]{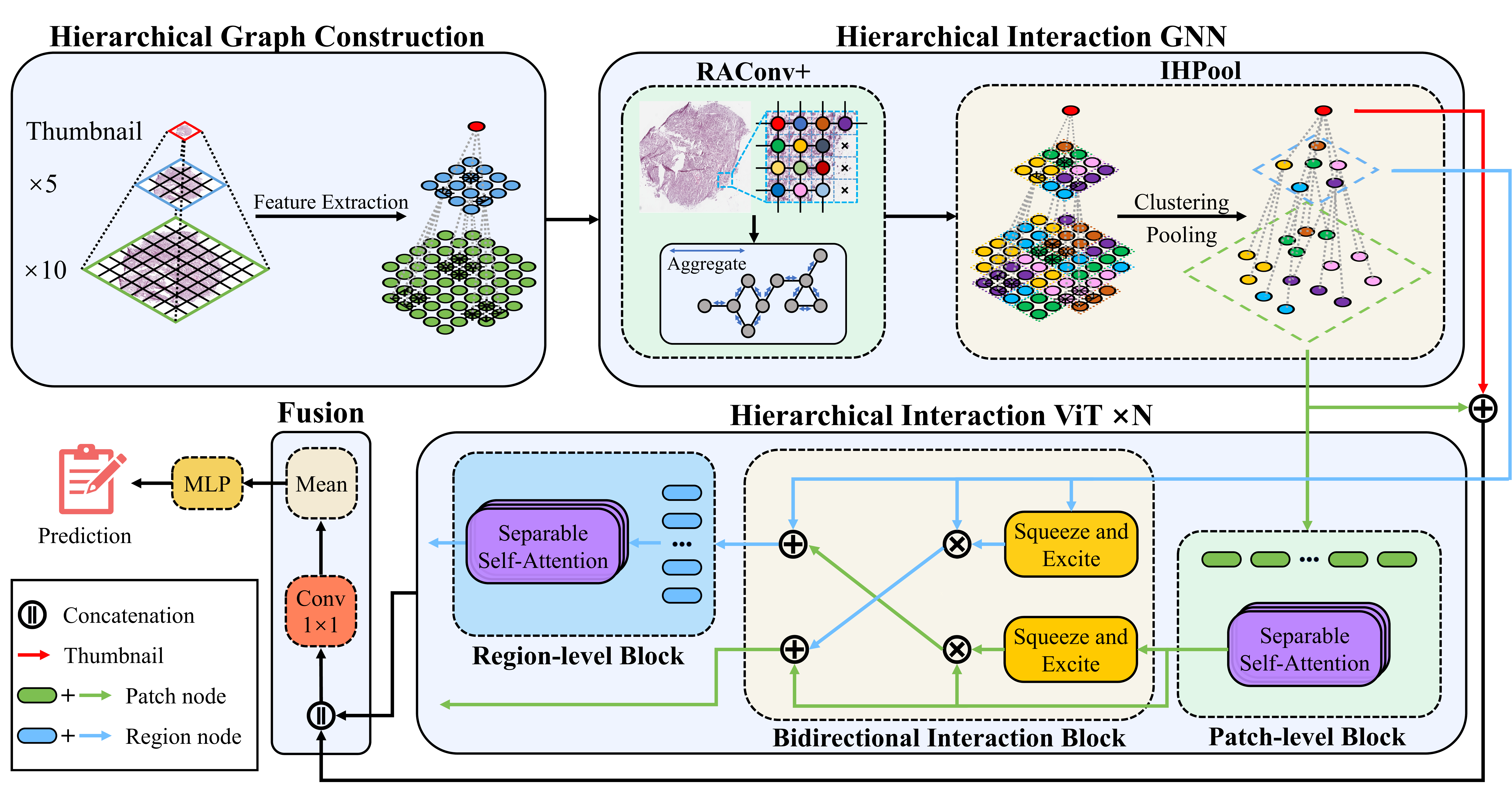}
\caption{Overview of the proposed HIGT framework. A WSI pyramid will be constructed as a hierarchical graph. Our proposed Hierarchical Interaction GNN and Hierarchical Interaction ViT block can
capture the local and global features, and the Bidirectional Interaction module in the latter allows the nodes from different levels to interact. And finally, the Fusion block aggregates the coarse-grained and fine-grained features to generate the slide-level prediction.} 
\label{overview}
\vspace{-0.3cm}
\end{figure}

Histopathology is considered the gold standard for diagnosing and treating many cancers~\cite{yao2020pathological}.
The tissue slices are usually scanned into Whole Slide Images (WSIs) and serve as important references for pathologists. 
Unlike natural images, WSIs typically contain billions of pixels and also have a pyramid structure, as shown in Fig.~\ref{overview}.
Such gigapixel resolution and expensive pixel-wise annotation efforts pose unique challenges to constructing effective and accurate models for WSI analysis. 
To overcome these challenges, Multiple Instance Learning (MIL) has become a popular paradigm for WSI analysis. 
Typically, MIL-based WSI analysis methods have three steps: (1) crop the huge WSI into numerous image patches; (2) extract instance features from the cropped patches; and (3) aggregate instance features together to obtain slide-level prediction results.
Many advanced MIL models emerged in the past few years.
For instance, ABMIL~\cite{ilse2018attention} and DeepAttnMIL~\cite{yao2020whole} incorporated attention mechanisms into the aggregation step and achieved promising results. 
Recently, Graph-Transformer architecture~\cite{wu2021representing} has been proposed to learn short-range local features through GNN and long-range global features through Transformer simultaneously.
Such Graph-Transformer architecture has also been introduced into WSI analysis~\cite{reisenbuchler2022local,zheng2022graph} to mine the thorough global and local correlations between different image patches. 
However, current Graph-Transformer-based WSI analysis models only consider the representation learning under one specific magnification, thus ignoring the rich multi-resolution information from the WSI pyramids. 

Different resolution levels in the WSI pyramids contain different and complementary information~\cite{chen2022scaling}.
The images at a high-resolution level contain cellular-level information, such as the nucleus and chromatin morphology features~\cite{kumar2017dataset}.
At a low-resolution level, tissue-related information like the extent of tumor-immune localization can be found~\cite{abduljabbar2020geospatial}, while the whole WSI describes the entire tissue microenvironment, such as intra-tumoral heterogeneity and tumor invasion~\cite{chen2022scaling}.
Therefore, analyzing from only a single resolution would lead to an incomplete picture of WSIs.
Some very recent works proposed to characterize and analyze WSIs in a pyramidal structure.
H2-MIL~\cite{hou2022h} formulated WSI as a hierarchical heterogeneous graph and HIPT~\cite{chen2022scaling} proposed an inheritable ViT framework to model WSI at different resolutions. 
Whereas these methods only characterize local or global correlations within the WSI pyramids and use only unidirectional interaction between different resolutions, leading to insufficient capability to model the rich multi-resolution information of the WSI pyramids. 

In this paper, we present a novel Hierarchical Interaction Graph-Transformer framework (\ie, HIGT) to simultaneously capture both local and global information from WSI pyramids with a novel Bidirectional Interaction module. 
Specifically, we abstract the multi-resolution WSI pyramid as a heterogeneous hierarchical graph and devise a Hierarchical Interaction Graph-Transformer architecture to learn both short-range and long-range correlations among different image patches within different resolutions.
Considering that the information from different resolutions is complementary and can benefit each other, we specially design a Bidirectional Interaction block in our Hierarchical Interaction ViT module to establish communication between different resolution levels.
Moreover, a Fusion block is proposed to aggregate features learned from the different levels for slide-level prediction.
To reduce the tremendous computation and memory cost, we further adopt the efficient pooling operation after the hierarchical GNN part to reduce the number of tokens and introduce the Separable Self-Attention Mechanism in Hierarchical Interaction ViT modules to reduce the computation burden.
The extensive experiments with promising results on two public WSI datasets from TCGA projects, \ie,  
kidney carcinoma (KICA) and esophageal carcinoma (ESCA), validate the effectiveness and efficiency of our framework on both tumor subtyping and staging tasks.
The codes are available at \href{https://github.com/mateguo0/HIGT}{https://github.com/HKU-MedAI/HIGT}.

\section{Methodology}
Fig.~\ref{overview} depicts the pipeline of HIGT framework for better exploring the multi-scale information in hierarchical WSI pyramids. 
First, we abstract each WSI as a hierarchical graph, where the feature embeddings extracted from multi-resolution patches serve as nodes and the edge denotes the spatial and scaling relationships of patches within and across different resolution levels.
Then, we feed the constructed graph into several hierarchical graph convolution blocks to learn the short-range relationship among graph nodes, following pooling operations to aggregate local context and reduce the number of nodes.
We further devise a Separable Self-Attention-based Hierarchical Interaction Transformer architecture equipped with a novel Bidirectional Interaction block to learn the long-range relationship among graph nodes. 
Finally, we design a fusion block to aggregate the features learned from the different levels of WSI pyramids for final slide-level prediction. 

\subsection{Graph Construction}
As shown in Fig.~\ref{overview}, a WSI is cropped into numerous non-overlapping $512\times512$ image patches under different magnifications (\ie, $\times5$, $\times10$) by using a sliding window strategy, where the OTSU algorithm~\cite{chen2021whole} is used to filter out the background patches.
Afterwards, we employ a pre-trained KimiaNet~\cite{riasatian2021fine} to extract the feature embedding of each image patch.
The feature embeddings of the slide-level $\boldsymbol{T}$ (Thumbnail), region-level $\boldsymbol{R}$ ($\times5$), and the patch-level $\boldsymbol{P}$ ($\times10$) can be represented as,
\begin{align}
\boldsymbol{T}&=\{\boldsymbol{t}\}, \nonumber \\
\boldsymbol{R}&=\{\boldsymbol{r}_1, \boldsymbol{r}_2, \cdots, \boldsymbol{r}_N\}, \nonumber \\
\boldsymbol{P} &= \{\boldsymbol{P}_{1}, \boldsymbol{P}_{2}, \cdots, \boldsymbol{P}_{N}\}, \boldsymbol{P_i} = \{\boldsymbol{p}_{i,1}, \boldsymbol{p}_{i,2}, \cdots, \boldsymbol{p}_{i,M}\}, 
\end{align}
where $\boldsymbol{t}, \boldsymbol{r}_i, \boldsymbol{p}_{i, j} \in \mathbb{R}^{1 \times C}$ correspond to the feature embeddings of each patch in thumbnail, region, and patch levels, respectively. 
$N$ is the total number of the region nodes and $M$ is the number of patch nodes belonging to a certain region node, and $C$ denotes the dimension of feature embedding (1,024 in our experiments). 
Based on the extracted feature embeddings, we construct a hierarchical graph to characterize the WSI, following previous H$^2$-MIL work~\cite{hou2022h}.
Specifically, the cropped patches serve as the nodes of the graph and we employ the extracted feature embedding as the node embeddings.
There are two kinds of edges in the graph: spatial edges to denote the 8-adjacent spatial relationships among different patches in the same levels, and scaling edges to denote the relationship between patches across different levels at the same location.

\subsection{Hierarchical Graph Neural Network}
To learn the short-range relationship among different patches within the WSI pyramid, we propose a new hierarchical graph message propagation operation, called RAConv+.
Specifically, for any source node $j$ in the hierarchical graph, we define the set of it all neighboring nodes at resolution $k$ as $\mathcal{N}_k$ and $k\in K$. Here $K$ means all resolutions. And the $h_k$ is the mean embedding of the node $j$'s neighboring nodes in resolution $k$. And $h_{j\prime}$ is the embedding of the neighboring nodes of node $j$ in resolution $k$ and ${h}_{{j\prime}} \in \mathcal{N}_k$.
The formula for calculating the attention score of node $j$ in resolution-level and node-level:
\begin{align}
\centering
&\alpha_k=\frac{\exp \left(\boldsymbol{a}^{\top} \cdot\operatorname{LeakyReLU}\left(\left[\boldsymbol{U} \boldsymbol{h}_j \| \boldsymbol{U} \boldsymbol{h}_k\right]\right)\right)}{\sum_{k^{\prime} \in \mathcal{K}} \exp \left(\boldsymbol{a}^{\top} \cdot\operatorname{LeakyReLU}\left(\left[\boldsymbol{U} \boldsymbol{h}_j \| \boldsymbol{U} \boldsymbol{h}_{k^{\prime}}\right]\right)\right)}, \nonumber \\
&\alpha_{{j\prime}}=\frac{\exp \left(\boldsymbol{b}^{\top} \cdot\operatorname{LeakyReLU}\left(\left[\boldsymbol{V} \boldsymbol{h}_j \| \boldsymbol{V} \boldsymbol{h}_{{j\prime}}\right]\right)\right)}{\sum_{
{h}_{{j\prime\prime}}
\in \mathcal{N}_k} \exp \left(\boldsymbol{b}^{\top} \cdot\operatorname{LeakyReLU}\left(\left[\boldsymbol{V} \boldsymbol{h}_j \| \boldsymbol{V} \boldsymbol{h}_{{j\prime\prime}}\right]\right)\right)}, \nonumber \\
& \alpha_{j,j\prime}=\alpha_k+ \alpha_{j\prime},
\label{eq3}
\end{align}%
 where $\alpha_{j,j\prime}$ is the attention score of the node $j$ to node $j\prime$ and $h_j$ is the source node $j$ embedding.
 And $U$, $V$, $a$ and $b$ are four learnable layers. The main difference from H2-MIL~\cite{HOU2021102092} is that we pose the non-linear $LeakyReLU$ between $a$ and $U$, $b$ and $V$, to generate a more distinct attention score matrix which increases the feature differences between different types of nodes~\cite{brody2021attentive}.
Therefore, the layer-wise graph message propagation can be represented as:
\begin{equation}
H^{(l+1)}=\sigma\left(\mathcal{A} \cdot H^{(l)} \cdot W^{(l)}\right),
\label{ceq1}
\end{equation}
where $\mathcal{A}$ represents the attention score matrix, and the attention score for the j-th row and j$\prime$-th column of the matrix is given by Eq.~\eqref{eq3}. At the end of the hierarchical GNN part, we use the IHPool\cite{HOU2021102092} progressively aggregate the hierarchical graph.

\subsection{Hierarchical Interaction ViT}
We further propose a Hierarchical Interaction ViT (HIViT) to learn long-range correlation within the WSI pyramids, which includes three key components: Patch-level (PL) blocks, Bidirectional Interaction (BI) blocks, and Region-level (RL) blocks. 

\para{Patch-level Block.} 
Given the patch-level feature set 
$\boldsymbol{P}=\bigcup_{i=1}^N \boldsymbol{P}_i$, 
the PL block learns long-term relationships within the patch level:
\begin{align}
    \hat{\boldsymbol{P}}^{l+1}=PL(\boldsymbol{P}^l)
\end{align}
where $l = 1, 2, ..., L$ is the index of the HIViT block. 
$PL(\cdot)$ includes a Separable Self Attention (SSA) \cite{mehta2022separable}, 1$\times$1 Convolution, and Layer Normalization in sequence.
Note that here we introduced SSA into the PL block to reduce the computation complexity of attention calculation from quadratic to linear while maintaining the performance\cite{mehta2022separable}. 

\para{Bidirectional Interaction Block.} 
We propose a Bidirectional Interaction (BI) block to establish communication between different levels within the WSI pyramids.
The BI block performs bidirectional interaction, and the interaction progress from region nodes to patch nodes is:
\begin{align}
& \boldsymbol{r_i^{l'}} \in \boldsymbol{R^{l'}}, \quad \boldsymbol{R^{l'}} = SE(\boldsymbol{R^l})\cdot \boldsymbol{R^l}, \nonumber \\
& \boldsymbol{P}_i^{l+1} = \{\boldsymbol{p}_{i,1}^{l+1},\boldsymbol{p}_{i,2}^{l+1}, \cdots , \boldsymbol{p}_{i,k}^{l+1}\},\quad \boldsymbol{p}_{i,k}^{l+1} = \hat{\boldsymbol{p}}_{i,k}^{l+1}+\boldsymbol{r}_i^{l'},  
\end{align}
where the $SE(\cdot)$ means the Sequeeze-and-Excite layer\cite{hu2018squeeze} and the $\boldsymbol{r_i^{l'}}$ means the i-th region node in $\boldsymbol{R^{l'}}$, and $\hat{\boldsymbol{p}}_{i,k}^{l+1}$ is the k-th patch node linked to the i-th region node after the interaction. Besides, another direction of the interaction is,
\begin{align}
&\bar{\boldsymbol{P}}^{l+1}  = \{\bar{\boldsymbol{P}}_1^{l+1},\bar{\boldsymbol{P}}_2^{l+1},\cdots, \bar{\boldsymbol{P}}_n^{l+1}\}, \quad\bar{\boldsymbol{P}}_i^{l+1} = MEAN(\hat{\boldsymbol{P}}_i^{l+1})\nonumber \\
& \hat{\boldsymbol{R}}^{l+1} = SE(\bar{\boldsymbol{P}}^{l+1} )\cdot \bar{\boldsymbol{P}}^{l+1}+\boldsymbol{R}^{l},  
\end{align}
where the $MEAN(\cdot)$ is the operation to get the mean value of patch nodes set $\hat{\boldsymbol{P}}_i^{l+1}$ associated with the i-th region node and $\bar{\boldsymbol{P}}_1^{l+1} \in \mathcal{R}^{1\times C}$ and the C is the feature channel of nodes, and $\hat{\boldsymbol{R}}^{l+1}$ is the region nodes set after interaction.

\para{Region-level Block.} 
The final part of this module is  to learn the long-range correlations of the interacted region-level nodes:
\begin{equation}
{\boldsymbol{R}}^{l+1}=RL(\hat{\boldsymbol{R}}^{l+1})
\end{equation}
where $l = 1, 2, ..., L$ is the index of the HIViT module, $\boldsymbol{R}=\{\boldsymbol{r}_1, \boldsymbol{r}_2, \cdots, \boldsymbol{r}_N\}$, and $RL(\cdot)$ has a similar structure to $PL(\cdot)$.

\subsection{Slide-level Prediction}
In the final stage of our framework, we design a Fusion block to combine the coarse-grained and fine-grained features learned from the WSI pyramids.
Specifically, we use an element-wise summation operation to fuse the coarse-grained thumbnail feature and patch-level features from the Hierarchical Interaction GNN part, and then further fuse the fine-grained patch-level features from the HIViT part with a concatenation operation.
Finally,  a $1\times1$ convolution and mean operation followed by a linear projection are employed to produce the slide-level prediction.
\section{Experiments}
\para{Datasets and Evaluation Metrics.}
We assess the efficacy of the proposed HIGT framework by testing it on two publicly available datasets (KICA and ESCA) from The Cancer Genome Atlas (TCGA) repository. 
The datasets are described below in more detail:
\begin{itemize}
    \item \textbf{KICA dataset.} The KICA dataset consists of 371 cases of kidney carcinoma, of which 279 are classified as early-stage and 92 as late-stage. 
    For the tumor typing task, 259 cases are diagnosed as kidney renal papillary cell carcinoma, while 112 cases are diagnosed as kidney chromophobe.
    \item \textbf{ESCA dataset.} The ESCA dataset comprises 161 cases of esophageal carcinoma, with 96 cases classified as early-stage and 65 as late-stage. 
    For the tumor typing task, there are 67 squamous cell carcinoma cases and 94 adenocarcinoma cases.
\end{itemize}

\para{Experimental Setup.} 
The proposed framework was implemented by PyTorch~\cite{paszke2019pytorch} and PyTorch Geometric~\cite{fey2019fast}.
All experiments were conducted on a workstation with eight NVIDIA GeForce RTX 3090 (24 GB) GPUs. 
The shape of all nodes' features extracted by KimiaNet is set to $1\times1024$.
All methods are trained with a batch size of 8 for 50 epochs.
The learning rate was set as 0.0005, with Adam optimizer.
The accuracy (ACC) and area under the curve (AUC) are used as the evaluation metric. 
All approaches were evaluated with five-fold cross-validations (5-fold CVs) from five different initializations.

\begin{table}[t]
\centering
\caption{Comparison with other methods on ESCA. Top results are shown in bold.
}
\begin{tabular}{m{2.60cm}<{\centering}m{2.20cm}<{\centering}m{2.20cm}<{\centering}m{2.20cm}<{\centering}m{2.20cm}<{\centering}}
\toprule[1pt]
\multirow{2}{*}{{Method}} &
\multicolumn{2}{c}{{Staging}} & \multicolumn{2}{c}{{Typing}} \\ 
\cmidrule(r){2-3}\cmidrule(r){4-5} & {AUC} & {ACC} & {AUC} & {ACC} \\ 
\bottomrule[1pt]

ABMIL~\cite{ilse2018attention} &
$64.53 \pm 4.80$ & $64.39 \pm 5.05$ &
$94.11 \pm 2.69$ & $93.07 \pm 2.68$  
 \\

CLAM-SB~\cite{lu2021data} &
$67.45 \pm 5.40$ & $67.29 \pm 5.18$ &
$93.79 \pm 5.52$ & $93.47 \pm 5.77$  
  \\

DeepAttnMIL~\cite{yao2020whole} &
$67.96 \pm 5.52$ & $67.53 \pm 4.96$ &
$95.68 \pm 1.94$ & $94.43 \pm 3.04$  
  \\

DS-MIL~\cite{li2021dual} &
$66.92 \pm 5.28$ & $66.83 \pm 5.57$ &
$95.96 \pm 3.07$ &$94.77 \pm 4.10$ 
   \\

LA-MIL~\cite{reisenbuchler2022local} &
$63.93 \pm 6.19$ & $63.45 \pm 6.19$ &
$95.23 \pm 3.75$ & $94.69 \pm 3.94$  
 \\

\cdashline{1-5}[2.0pt/2.0pt]

H2-MIL~\cite{hou2022h} &
$63.20\pm8.36$  & $62.72\pm8.32$ &
$91.88\pm4.17$ & $91.31\pm4.18$   
  \\

HIPT~\cite{chen2022scaling} &
$68.59 \pm 5.62$ & $68.45 \pm 6.39$ &
$94.62 \pm 2.34$ & $93.01 \pm 3.28$  
  \\

\bottomrule[1pt]

\textbf{Ours} &
$\textbf{71.11}\pm\textbf{6.04}$ & $\textbf{70.53}\pm\textbf{5.41}$ &
$\textbf{96.81}\pm\textbf{2.49}$ & $\textbf{96.16}\pm\textbf{2.85}$    \\
\bottomrule[1pt]
\end{tabular}
\label{Comparison with SOTAs on ESCA Dataset}
\end{table}

\begin{table}[t]
\centering
\caption{Comparison with other methods on KICA. Top results are shown in bold.
}
\begin{tabular}{m{2.60cm}<{\centering}m{2.20cm}<{\centering}m{2.20cm}<{\centering}m{2.20cm}<{\centering}m{2.20cm}<{\centering}}
\toprule[1pt]
\multirow{2}{*}{{Method}} &
\multicolumn{2}{c}{{Staging}} & \multicolumn{2}{c}{{Typing}} \\ 
\cmidrule(r){2-3}\cmidrule(r){4-5} & {AUC} & {ACC} & {AUC} & {ACC} \\ 
\bottomrule[1pt]

ABMIL~\cite{ilse2018attention} &
$77.40 \pm 3.87$  & $75.94 \pm 5.06$ &
$97.76 \pm 1.74$ &  $98.86 \pm 0.69$ 
 \\

CLAM-SB~\cite{lu2021data} &
$77.16 \pm 3.64$ & $76.61 \pm 4.31$ &
$96.76 \pm 3.42$ & $97.13 \pm 2.99$ 
 \\

DeepAttnMIL~\cite{yao2020whole} &
$76.77 \pm 1.94$ & $75.94 \pm 2.41$ &
$97.44 \pm 1.04$ & $96.30 \pm 2.63$  
  \\

DS-MIL~\cite{li2021dual} &
$77.33 \pm 4.11$ & $76.57 \pm 5.14$ &
$98.03 \pm 1.13$ & $97.31 \pm 1.85$  
   \\

LA-MIL~\cite{reisenbuchler2022local} &
$69.37 \pm 5.27$ & $68.73 \pm 5.09$ &
$98.34 \pm 0.98$ & $97.71 \pm 1.76$  
  \\

\cdashline{1-5}[2.0pt/2.0pt]

H2-MIL~\cite{hou2022h} &
$65.59 \pm 6.65$ & $64.48 \pm 6.20$ &
$98.06 \pm 1.43$ & $96.99 \pm 3.01$ 
 \\

HIPT~\cite{chen2022scaling} &
$75.93 \pm 2.01$ & $75.34 \pm 2.31$ &
$98.71 \pm 0.49$ & $97.32 \pm 2.24$ 
 \\

\bottomrule[1pt]

\textbf{Ours} &
$\textbf{78.80}\pm\textbf{2.10}$ & $\textbf{76.80}\pm\textbf{2.30}$ &
$\textbf{98.90}\pm\textbf{0.60}$ & $\textbf{97.90}\pm\textbf{1.40}$     \\
\bottomrule[1pt]
\end{tabular}
\label{Comparison with SOTAs on KICA Dataset}
\end{table}

\para{Comparison with State-of-the-art Methods.}
We first compared our proposed HIGT framework with two groups of state-of-the-art WSI analysis methods: 
(1) non-hierarchical methods including: 
ABMIL~\cite{ilse2018attention}, 
CLAM-SB~\cite{lu2021data}, 
DeepAttnMIL~\cite{yao2020whole}, 
DS-MIL~\cite{li2021dual}, 
LA-MIL~\cite{reisenbuchler2022local}, 
and 
(2) hierarchical methods including:
H2-MIL~\cite{hou2022h}, 
HIPT~\cite{chen2022scaling}. 
For LA-MIL~\cite{reisenbuchler2022local} method, it was introduced with a single-scale Graph-Transformer architecture.
For H2-MIL~\cite{hou2022h} and HIPT~\cite{chen2022scaling}, they were introduced with a hierarchical Graph Neural Network and hierarchical Transformer architecture, respectively.
The results for ESCA and KICA datasets are summarized in Table~\ref{Comparison with SOTAs on ESCA Dataset} and Table~\ref{Comparison with SOTAs on KICA Dataset}, respectively.
Overall, our model achieves a content result both in AUC and ACC of classifying the WSI, and especially in predicting the more complex task (i.e. Staging) compared with the SOTA approaches. 
Even for the non-hierarchical Graph-Transformer baseline LA-MIL and hierarchical transformer model HIPT, our model approaches at least around 3\% and 2\% improvement on AUC and ACC in the classification of the Staging of the KICA dataset. Therefore we believe that our model benefits a lot from its used modules and mechanisms.

\para{Ablation Analysis.}
We further conduct an ablation study to demonstrate the effectiveness of the proposed components. The results are shown in Table~\ref{Ablation study on KICA dataset}. 
In its first row, we replace the RAConv+ with the original version of this operation. And in the second row, we replace the Separable Self Attention with a canonical transformer block. The third row changes the bidirectional interaction mechanism into just one direction from region-level to patch-level. And the last row, we remove the fusion block from our model. Finally, the ablation analysis results show that all of these modules we used actually improved the prediction effect of the model to a certain extent.

\begin{figure}[t]
\includegraphics[width=\textwidth]{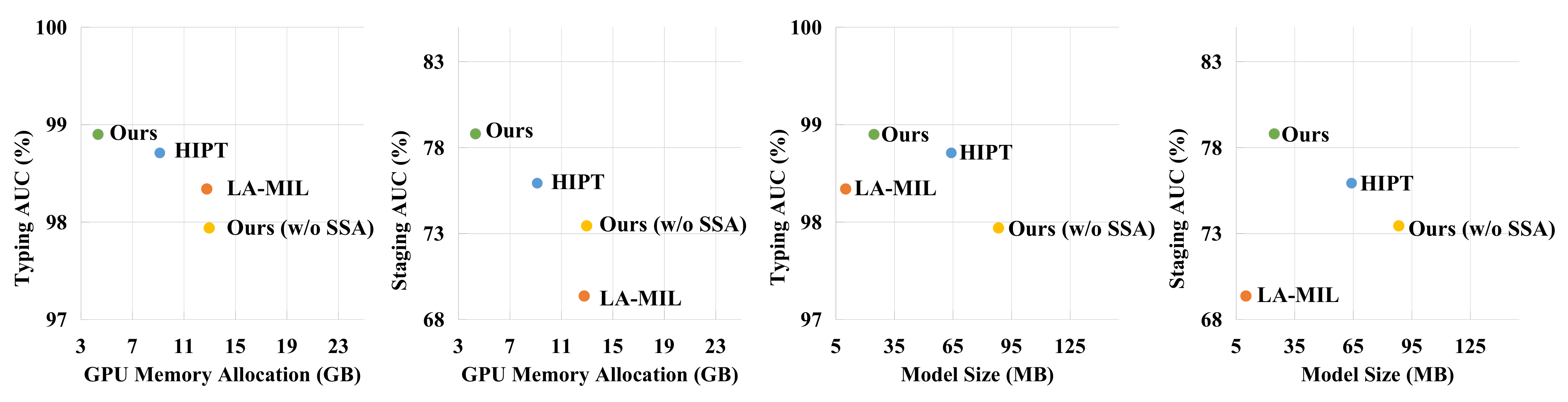}
\caption{Computational analysis of our framework and some selected SOTA methods. From left to right are scatter plots of Typing AUC v.s. GPU Memory Allocation, Staging AUC v.s. GPU Memory Allocation, Typing AUC v.s. Model Size, Staging AUC v.s. Model Size.}
\label{computation}
\end{figure}

\para{Computation Cost Analysis.} 
We analyze the computation cost during the experiments to compare the efficiency between our methods and existing state-of-the-art approaches. 
Besides we visualized the model size (MB) and the training memory allocation of GPU (GB) v.s. performance in KICA's typing and staging task plots in Fig.~\ref{computation}.
All results demonstrate that our model is able to maintain the promising prediction result while reducing the computational cost and model size effectively.

\begin{table}[t]
\centering
\caption{Ablation analysis on KICA dataset.
}
\begin{tabular}{m{2.60cm}<{\centering}m{2.20cm}<{\centering}m{2.20cm}<{\centering}m{2.20cm}<{\centering}m{2.20cm}<{\centering}}
\toprule[1pt]
\multirow{2}{*}{{Method}} &
\multicolumn{2}{c}{{Staging}} & \multicolumn{2}{c}{{Typing}} \\ 
\cmidrule(r){2-3}\cmidrule(r){4-5} & {AUC} & {ACC} & {AUC} & {ACC} \\ 
\bottomrule[1pt]
H2-MIL $+$ HIViT &
$77.35 \pm 3.41$ & $\textbf{77.16} \pm \textbf{3.29}$ &
$98.56 \pm 1.01$ & $95.00 \pm 1.75$ \\

Ours w/o SSA &
$73.45 \pm 8.48$ & $71.47 \pm 3.21$ &
$97.94 \pm 2.51$ & $97.42 \pm 2.65$ \\

Ours w/o BI &
$72.42 \pm 2.09$ & $71.34 \pm 7.23$ &
$98.04 \pm 8.30$ & $96.54 \pm 2.80$ \\

Ours w/o Fusion &
$77.87 \pm 2.09$ & $76.80 \pm 2.95$ &
$98.46 \pm 0.88$ & $97.35 \pm 1.81$ \\

\bottomrule[1pt]

\textbf{Ours} &
$\textbf{78.80}\pm\textbf{2.10}$ & $76.80\pm2.30$ &
$\textbf{98.90}\pm\textbf{0.60}$ & $\textbf{97.90}\pm\textbf{1.40}$     \\
\bottomrule[1pt]
\end{tabular}
\label{Ablation study on KICA dataset}
\end{table}

\section{Conclusion}
In this paper, we propose HIGT, a framework that simultaneously and effectively captures local and global information from the hierarchical WSI.
Firstly, the constructed hierarchical data structure of the multi-resolution WSI is able to offer multi-scale information to the later model. 
Moreover, the redesigned H2-MIL and HIViT capture the short-range and long-range correlations among varying magnifications of WSI separately. 
And the bidirectional interaction mechanism and fusion block can facilitate communication between different levels in the Transformer part. 
We use IHPool and apply the Separable Self Attention to deal with the inherently high computational cost of the Graph-Transformer model.
Extensive experimentation on two public WSI datasets demonstrates the effectiveness and efficiency of our designed framework, yielding promising results. 
In the future, we will evaluate on other complex tasks such as survival prediction and investigate other techniques to improve the efficiency of our framework.

\section*{Acknowledgement} The work described in this paper was partially supported by grants from the National Natural Science Fund (62201483), the Research Grants Council of the Hong Kong Special Administrative Region, China (T45-401/22-N), and The Hong Kong Polytechnic University (P0045999).

\clearpage
\bibliographystyle{splncs04}
\bibliography{reference}

%




\end{document}